\documentclass[runningheads]{llncs}
\usepackage{graphicx}

\usepackage[hidelinks,colorlinks,allcolors=blue]{hyperref}
\usepackage{booktabs}
\usepackage{threeparttable}
\usepackage{colortbl}
\usepackage{amsmath,amssymb}

\usepackage{cleveref}
\Crefname{figure}{Fig.}{Figs.}
\Crefname{section}{Sect.}{Sects.}

\definecolor{mygray}{gray}{0.9}

\begin{document}
\title{Pavlok-Nudge: A Feedback Mechanism for Atomic Behaviour Modification with Snoring Usecase%
\thanks{Md Rakibul Hasan and Shreya Ghosh are co-first authors.}%
}
\titlerunning{Pavlok-Nudge: A Feedback Mechanism for Atomic Behaviour Modification}
%
\author{Md Rakibul Hasan\inst{1}\orcidID{0000-0003-2565-5321} \and
Shreya Ghosh\inst{1}\orcidID{0000-0002-2639-8374} \and
Pradyumna Agrawal\inst{1} \and
Zhixi Cai\inst{2}\orcidID{0000-0001-7978-0860} \and
Abhinav Dhall\inst{2}\orcidID{0000-0002-2230-1440} \and
Tom Gedeon\inst{1,3,4}\orcidID{0000-0001-8356-4909}}
\authorrunning{M. R. Hasan et al.}

\institute{School of Electrical Engineering, Computing and Mathematical Sciences, Curtin University, Bentley, WA 6102, Australia \and
Faculty of Information Technology, Monash University, Clayton, VIC 3800, Australia \and
The Australian National University, Canberra ACT 2600, Australia \and
Obuda University, Budapest, Hungary \\
\email{\{Rakibul.Hasan, Shreya.Ghosh, Tom.Gedeon\}@curtin.edu.au \\
\{Zhixi.Cai, Abhinav.Dhall\}@monash.edu}}
%
\maketitle              
\begin{abstract}
This paper proposes an atomic behaviour intervention strategy using the Pavlok wearable device. Pavlok utilises beeps, vibration and shocks as a mode of aversion technique to help individuals with behaviour modification. While the device can be useful in certain periodic daily life situations, like alarms and exercise notifications, it relies on manual operations that limit its usage. To automate behaviour modification, we propose a framework that first detects targeted behaviours through a lightweight deep learning model and subsequently nudges the user. Our proposed solution is implemented and verified in the context of snoring, which captures audio from the environment following a prediction of whether the audio content is a snore or not using a \emph{lightweight} 1D convolutional neural network. Based on the prediction, we use Pavlok to nudge users for preventive measures, such as a change in sleeping posture. We believe that this simple solution can help people change their atomic habits, which may lead to long-term health benefits. Our proposed lightweight model (99.8\% fewer parameters over SOTA; 790,273$\rightarrow$1,337) achieves SOTA test accuracy of 0.99 on a public benchmark. The code and model are publicly available at \url{https://github.com/hasan-rakibul/pavlok-nudge-snore}.
\end{abstract}

\section{Introduction}\label{sec:introduction}
Software-based \textit{behavioural intervention technologies (BITs)} have the potential to transform human atomic behaviour~\cite{schueller2013realizing}, which could further lead to long-term health benefits that directly impact human cognition and emotional states~\cite{saha2017affective}. The emerging field of BITs in pervasive computing mainly includes websites, mobile applications and internet of things (IoT) sensor-based designs aiming to make people's lives better~\cite{englhardt2023classification,liu2024rppg}. To this end, we leverage the Pavlok device, a wearable aimed to help people ``break bad habits'' using aversion therapy~\cite{pavlokapi}. It accomplishes this by providing nudges in the form of \textit{beeps}, \textit{vibration} and/or \textit{electric shock} to the wearer's wrist when they engage in targeted behaviours, for example, snoring. The main intention behind this approach is that over time, the wearer is expected to become conscious of the atomic behaviour due to the unpleasant sensation of the beep/vibration/shock, which may help the individual to be less inclined to repeat the behaviour. Moreover, Pavlok can also work in synchronisation with a smartphone application to monitor progress by setting up behaviour change goals. While the native Pavlok is generally useful in certain periodic daily life situations, like alarms, exercise notifications and nail-biting, the device mostly relies on manual operation by the user rather than automatic action detection, which limits its usefulness. Similarly, it lacks the advanced sensory capabilities to detect and take action on relevant events in real time.

\begin{figure}[t]
    \centering
    \includegraphics[width=1\linewidth]{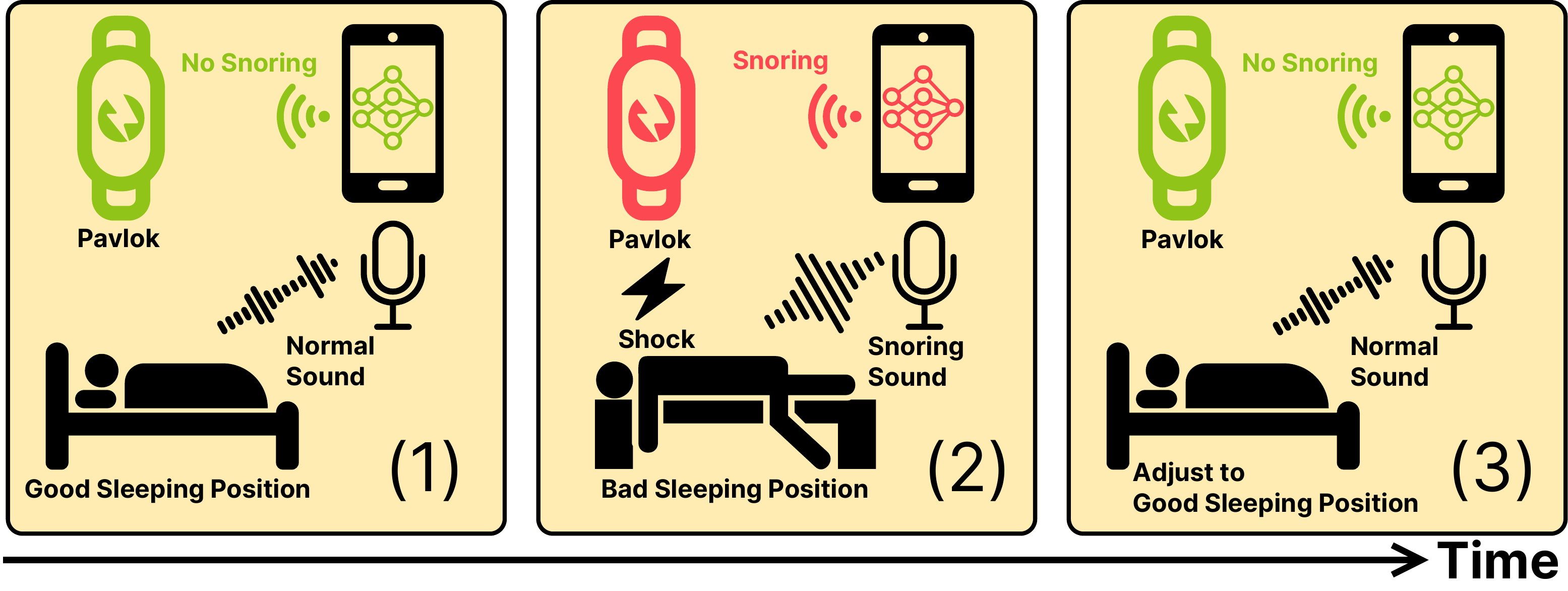}
    \caption{\small \underline{Pavlok-Nudge Overview.} Pavlok-Nudge is an interactive, nearly real-time software application aiming to aid individuals in atomic behaviour modification. Here, we present the use case for snoring behaviour modification. Our proposed solution consists of the wrist-borne Pavlok and a smartphone (see 1, 2, 3). The microphone on the mobile device captures audio, which is further passed through a deep learning-based framework to detect snoring sounds. Upon snoring detection, the mobile device provides a signal to the Pavlok to nudge the user for a posture change to stop snoring (see 2). Scenarios (1), (2) and (3) happen sequentially over time.}
    \label{fig:overview}
\end{figure}

\noindent \textbf{Motivation.} 
Snoring is a harsh sound produced during sleep when the throat muscles relax, narrowing the airway and causing it to vibrate as air flows through. According to medical research~\cite{yaremchuk2020why}, a natural way to reduce or avert snoring is to sleep on the side. However, maintaining this posture involuntarily could be difficult when a person sleeps. Our proposed system, \textit{Pavlok-Nudge}, can help individuals with behaviour training. Snoring is strongly correlated with obstructive sleep apnea~\cite{nguyen2022snore,alencar2013dynamics}, a condition that causes a person to stop breathing periodically. Sleep apnea causes several health issues such as heart disease, diabetes, obesity, arrhythmia (irregular heart rhythm), headache and stroke~\cite{hagg2022negative}. Poor quality of sleep reduces memory, thinking capability and proper mental functioning in daily life situations. Prolonged compromises in sleep can lead to frequent irritability, short temper, and even depression~\cite{hagg2022negative}. 

For our use case, a mobile phone's microphone is used to capture snoring noises (\Cref{fig:overview}). The phone can establish a Bluetooth low energy (BLE) connection with the Pavlok device, which receives a signal to deliver a ``nudge'' or ``shock'' if snoring is detected. This way, our proposed solution augments the sensory capabilities of the Pavlok. The proposed solution also allows greater customisation and control over Pavlok's responses related to context. Users can switch between beep, vibration and shock, along with the intensity of the nudge stimulus. Thus, overall, it is user-friendly and adjustable to best suit each individual's needs. The main contribution of the paper is as follows:
\begin{itemize}
    \item \textit{Pavlok-Nudge} is a software-based behaviour intervention technology to help individuals \textit{break behavioural patterns}. It is an interactive and open-sourced solution that will help the research community extend it to other human behaviour modification settings.
    
    \item To the best of our knowledge, we are the first to develop a nudge-based approach to snoring-based behaviour modification. On a dataset of 1,000 snoring and non-snoring sounds, our lightweight model (99.8\% less parameters over SOTA; 1,278,049 $\rightarrow$ 1,337) achieves a SOTA test accuracy of 0.99. 
    
\end{itemize}

\begin{table}[t]
    \centering
    \caption{\textbf{Where do we stand now?} Overview of prior studies on snore detection.}
    \label{tab:overview}
    \resizebox{\linewidth}{!}{%
    \begin{tabular}{@{}*{6}cl@{}}
    \toprule
    \rowcolor{mygray}
       \textbf{Ref.}  & \textbf{Year} & \textbf{Dataset Details} & \textbf{Data Avail.} & \textbf{Code Avail.} & \textbf{Feature} & \textbf{Classifier}   \\ \midrule
      \cite{nguyen2015sleep}   &  2015 & \# Sub: 15 & $\times$ & $\times$ & Raw & f-MLP \\
      \cite{ccavucsouglu2017acoustics}   &  2017 & \# Sub: 68, Lab & $\times$ & $\times$ & Frequency & MLP \\
      \cite{arsenali2018recurrent}   &  2018 & \# Sub: 20, Hospital & $\times$ & $\times$ & MFCC & RNN \\
      \cite{sun2019snorenet}   &  2019 & \# Sub: 10, Bedroom & $\times$ & $\times$ & Raw & 1D-CNN \\
      \cite{khan2019deep}   &  2019 & \# Sam: 1000, Web & \checkmark & $\times$ & MFCC & 2D-CNN \\
      \cite{jiang2020automatic}   &  2020 & \# Sub: 15, Hospital & $\times$ & $\times$ & Mel Spectrogram & CNN-LSTM-DNN \\
      \cite{xie2021audio}   &  2021 & \# Sub: 38, Lab & $\times$ & $\times$ & CQT Spectrogram & CNN-LSTM \\
      \cite{oran2022malefemale} & 2022 & \# Sam: 1000 & \checkmark & \checkmark & Mel Spectrogram & CNN \\
      \cite{li2023automatic}   &  2023 & \# Sub: 88, Home & $\times$ & $\times$ & Raw; Visibility Graph & 1D-2D-CNN \\ 
      \cite{dong2025multi} & 2025 & \cite{khan2019deep} & \checkmark & $\times$ & MFCC & Multi-branch CNN \\
      \cite{kiruthika2025novel} & 2025 & \cite{khan2019deep} & \checkmark & $\times$ & Mel Spectrogram & CNN-GRU \\
      \cite{pachori2025automated} & 2025 & \cite{khan2019deep} & \checkmark & $\times$ & Time-frequency & VGG19 \\
      \cite{pachori2025automated2} & 2025 & \cite{khan2019deep} & \checkmark & $\times$ & Time-frequency & VGG16-SVM \\
      \textbf{Our}   &  \textbf{2025} & \textbf{\cite{khan2019deep,oran2022malefemale}} & \textbf{\checkmark} & \textbf{\checkmark} & \textbf{MFCC} & \textbf{1D-CNN} \\
      \bottomrule
    \end{tabular}%
    }
\end{table}





\begin{figure}[!t]
    \centering
    \includegraphics[width=\textwidth]{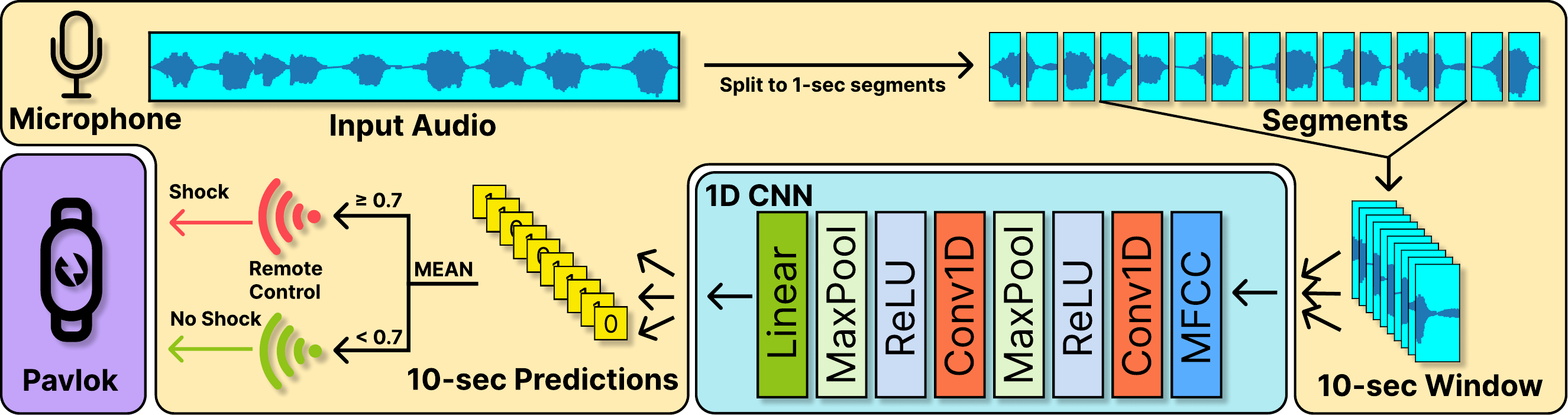}
    \caption{\small \underline{Pavlok-Nudge Workflow.} The Pavlok-Nudge framework senses the environment using the microphone of a mobile or laptop device. To process the audio input, a non-overlapping sliding window-based approach divides the input audio into 1-second chunks, which are then fed to a deep learning-based snore detection model. After collecting predictions on a certain number of chunks (e.g., 10 seconds), a voting strategy (e.g., 70\%) is used to determine if the incoming signal belongs to the snore category. Finally, upon reaching a decision, the system utilises Bluetooth to connect with the Pavlok device and deliver a user-specific ``nudge''. Refer to \Cref{sec:workflow} for additional details or model architecture, such as kernel size, dropout rate and activation functions.}
    \label{fig:workflow}
\end{figure}

\section{Related Work}
\Cref{tab:overview} presents an overview of prior work on snore detection using various deep learning methods. One of the earliest versions of snore recording and detection software was developed by SnoreLab~\cite{snorelab}. However, the interface does not provide a snore prevention strategy to act upon snore detection. Non-deep learning-based methods, such as a logistic regression classifier with auditory image modelling features~\cite{nonaka2016automatic} and an AdaBoost classifier with spectral features~\cite{dafna2013automatic}, were utilised for snore detection. Recently, Romero et al.~\cite{romero2019deep} and Emoto et al.~\cite{emoto2018detection} use latent space features from a pre-trained deep learning model to predict if the incoming audio signal is a snore. Similarly, Arsenali et al.~\cite{arsenali2018recurrent} use a recurrent neural network with mel-frequency cepstral coefficients (MFCC) features to detect snoring. The most closely related work by Khan~\cite{khan2019deep} also utilises a convolutional neural network (CNN)-based deep learning framework with MFCC features to detect snoring. The proposed solution also includes a vibration device to alert the user to change sleeping posture. However, their system relies on wired connections that may cause user discomfort, and its deep learning model is relatively large and computationally intensive. Our proposed solution is more user-friendly and lightweight for easier deployment.


\section{Pavlok-Nudge Pipeline}\label{sec:workflow}
Our proposed ``Pavlok-Nudge'' solution consists of a deep learning-based snore-detection model, mobile application and Pavlok (\Cref{fig:workflow}).

\subsection{Deep Learning-Based Snore Detection Model}
\subsubsection{Problem Formulation.}
Let the continuous audio signal captured by the mobile's microphone be denoted as $x(t)$, where $t$ represents time. The audio is segmented into non-overlapping 1-second chunks, $x_i$, such that:
\begin{equation}
x = \{x_1, x_2, \dots, x_N\}, \quad \text{where } i = 1, 2, \dots, N.
\end{equation}

Each 1-second audio segment $x_i \in \mathbb{R}$ is processed to extract MFCCs. As suggested in prior work \cite{hasan2021how}, we compute 25 MFCC components at 85 evenly spaced time points, resulting in a feature matrix $X_i \in \mathbb{R}^{25 \times 85}$ for each audio segment $x_i$. 

Thus, the input becomes a sequence of $X = \{X_1, X_2, \dots, X_N\}$, where $X_i$ serves as the representation of the corresponding audio segment $x_i$. Our goal is to develop a classifier $\mathcal{F}$ that maps each input audio segment $X_i \in \mathbb{R}^{25 \times 85}$ to a binary output $y_i \in \{0, 1\}$, where $y_i = 1$ indicates the presence of snoring and $y_i = 0$ otherwise.

\subsubsection{Model Architecture.}
Our proposed 1D CNN architecture captures local temporal patterns in the MFCC time series with a relatively small parameter footprint, which is crucial for deployment with limited computational resources. After evaluating various configurations of convolutional and fully connected layers, we finalised a model with only two convolutional layers, each using $3\times3$ kernels, followed by a fully connected layer. Each convolutional layer is followed by a \texttt{ReLU} activation function and max-pooling. The first convolutional layer outputs 8 channels, while the second expands this to 16 channels, allowing for progressively richer feature extraction from the input audio signal. Max-pooling layers with $2\times2$ kernels reduce the spatial dimensions by half. After the convolutional layers, the network output is flattened and passed through a fully connected layer. To mitigate overfitting, a dropout layer with a 0.4 rate is applied before the final classification layer. The output is then passed through a \texttt{Sigmoid} activation function to produce a probability for binary classification. This lightweight architecture effectively captures both temporal and frequency features of snore sounds while maintaining a compact footprint of just 1,337 trainable parameters.

\subsubsection{Evaluation Protocol.}
We report the performance of our model in terms of standard classification metrics, including classification accuracy, precision, recall, F1 score and area under the receiver operating characteristic curve (AUROC).

\subsection{Real-Time Monitoring} 
The detection cycle in our proposed system comprises continuous audio recording using the microphone of a device such as a smartphone or a computer. For our experiments, we use a 10-second recording window to balance timely detection with sufficient context. This audio clip is then segmented into 1-second intervals, which are processed through the pre-trained snore detection model $\mathcal{F}$. If a specified proportion of the audio segments (e.g., 70\%) is identified as containing snoring incidents, a pre-selected stimulus -- such as beeps, vibration or an electric shock -- is administered to the user via the Pavlok device, in accordance with their preferences. The detection cycle is designed to continuously monitor snoring behaviour and provide a prompt and effective aversive stimulus to facilitate behavioural modification.


\section{Experiments and Results}

\subsection{Dataset}
We utilise two snoring datasets. \textbf{(1) Khan}~\cite{khan2019deep} released a collection of 500 snoring and 500 non-snoring sounds obtained from various online sources. After removing the silent segments from the audio recordings, individual samples lasting one second each were separated. The snoring sounds were derived from individuals spanning different age groups, including children, adult men and adult women. Among the 500 snoring samples, 72.6\% were characterised by the absence of any background noise, while the remaining 27.4\% exhibited non-snoring background noise. As for the 500 non-snoring samples, the dataset consists of 50 samples for each of the 10 distinct ambient sounds commonly encountered near individuals during sleep, such as a baby crying, clock ticking, toilet flushing, sirens, television noise, car noise, people talking, rain and thunderstorms. Following~\cite{khan2019deep}, we split the dataset into three partitions: 700 samples for training, 200 samples for validation and the remaining 100 samples for testing the model. 

\textbf{(2) MaleFemale} dataset, collected from Kaggle~\cite{oran2022malefemale}, consists of 1,000 snoring sounds: 500 from males and 500 from females. As there are no non-snoring sounds in this dataset, we merged this dataset with the Khan \cite{khan2019deep} dataset, resulting in a dataset of 2,000 samples. Like Khan \cite{khan2019deep}, we randomly selected 100 samples from the MaleFemale dataset and made a consolidated test set of 200 samples (i.e., 100 from Khan and 100 from the MaleFemale dataset). This merged dataset includes 1,600 training samples (700 from Khan and 900 from the MaleFemale dataset).

\subsection{Implementation Details}
The experiments were conducted on a system running SUSE Linux Enterprise Server 15 SP4 with Python 3.12.3 as the programming language. We utilised \texttt{PyTorch} as the core framework, along with \texttt{torchaudio} for audio processing, \texttt{torchmetrics} for evaluation metrics, and \texttt{pytorch-lightning} for efficient model training. Additionally, \texttt{ffmpeg} was used to support audio data processing. We developed the software interface using JavaScript and Bluetooth technology to establish a connection with the Pavlok 3 device~\cite{pavlokapi}.

The CNN models are trained using the \texttt{Adam} optimiser with a learning rate of 0.001 to minimise the binary cross-entropy loss. Training is capped at a maximum of 100 epochs. We tuned various hyperparameters, including the learning rate, dropout rate and convolutional layer sizes, to optimise the architecture. To prevent overfitting and ensure optimal generalisation, we implement an early stopping criterion, halting the training if the validation loss does not improve for three consecutive epochs. The best-performing model, identified by the lowest validation loss during training, is saved to ensure the most effective version is retained for subsequent inference.

\subsection{Main Result}
The performance of our snore detection model across different configurations is presented in \Cref{tab:comp}. Our final model achieved an accuracy of 0.99, along with high precision (0.98), recall (1.00), F1 score (0.99) and AUROC (0.99), on the test set. While two recent works \cite{dong2025multi,kiruthika2025novel}, published in 2025, also reported competitive performances, their model size is much larger (parameter count of 39.9 and 0.79 million, respectively) with a more complex architecture. In addition to superior performance, our model maintains an exceptionally low parameter footprint of only 1,337 trainable parameters.

\begin{table}[!t]
    \centering
    \caption{Performance of our proposed model and its comparison with the literature.}
    \label{tab:comp}
    \resizebox{\linewidth}{!}{%
    \begin{threeparttable}
    \begin{tabular}{l*8c} \toprule
   \rowcolor{mygray}   \textbf{Dataset} & \textbf{Model} & \textbf{O/p Ch $\downarrow$} & \textbf{Params $\downarrow$} & \textbf{Acc $\uparrow$} & \textbf{P $\uparrow$} & \textbf{R $\uparrow$} & \textbf{F1 $\uparrow$} & \textbf{AUROC $\uparrow$} \\ \midrule
    Khan \cite{khan2019deep} & 2D CNN \cite{khan2019deep} & 32-32-64-64 & \textit{1.3 M} & 0.96 & -- & -- & -- & -- \\
        & Multi-branch CNN \cite{dong2025multi} & 384- $\cdots$ -2144 & \textit{39.9 M} & \textbf{0.99} & -- & \textbf{1.00} & \textbf{0.99} & -- \\ 
        & CNN-GRU \cite{kiruthika2025novel} & -- & 0.79 M & \textbf{0.99} & \textbf{0.99} & 0.99 & 0.99 & \textbf{0.99} \\
        & VGG19 \cite{pachori2025automated} & 64- $\cdots$ -512 & 144 M & 0.97 & 0.94 & 0.99 & 0.96 & \textbf{1.00} \\
        & VGG16-SVM \cite{pachori2025automated2} & 64- $\cdots$ -512 & 138 M & 0.96 & 0.96 & 0.96 & 0.96 & -- \\
        & 2D CNN\tnote{a} \cite{khan2019deep} & 32-32-64-64 & 1.3 M & 0.95 & 0.91 & \textbf{1.00} & 0.95 & 0.95 \\
            & \textbf{1D CNN} \textbf{(Ours)} & \textbf{8-16} & \textbf{1,337} & \textbf{0.99} & 0.98 & \textbf{1.00} & \textbf{0.99} & 0.99 \\ \midrule
        \parbox{7.5em}{\raggedright Khan \cite{khan2019deep} + MaleFemale \cite{oran2022malefemale}} 
                & \textbf{1D CNN} \textbf{(Ours)} & 8-16 & 1,337 & 0.96 & 0.97 & 0.98 & 0.97 & 0.94 \\
        \bottomrule
    \end{tabular}
    \begin{tablenotes}
        \item[a] Our implementation of \cite{khan2019deep}'s model, evaluated on \emph{identical} test samples as ours.
        \item ``--'' means \textit{unreported}.
        \item Parameters in \textit{italics} refer to our calculation based on the architectures reported in the corresponding papers.
        \item O/p Ch -- number of channels in the output of \emph{convolutional} layers
        \item Acc -- classification accuracy; Params -- number of trainable parameters; P -- precision; R -- recall
    \end{tablenotes}
    \end{threeparttable}%
    }
\end{table}

The pioneering work of Khan~\cite{khan2019deep} proposed an MFCC feature-based CNN architecture and reported a classification accuracy of 0.96. The Khan \cite{khan2019deep} dataset does not have any fixed, separated test samples. Therefore, to ensure a fairer comparison by evaluating on \emph{identical} test samples, we reimplemented Khan~\cite{khan2019deep}'s proposed model\footnote{Our implementation of \cite{khan2019deep} achieved a classification accuracy of 0.95, compared to the 0.96 accuracy reported by them. This slight difference is likely due to variations in data preprocessing methods.}. As shown in \Cref{tab:comp}, our proposed 1D CNN model outperforms \cite{khan2019deep}'s 2D CNN model in terms of accuracy, precision, F1 score and AUROC. Both Khan \cite{khan2019deep} and our model provide a recall of 1.00, indicating that it is easier to identify all snore instances in this dataset, with no false negatives.

In terms of the aversion technique, Khan \cite{khan2019deep} proposed a custom-made embedded system affixed to the upper arm, which uses a vibration-based feedback mechanism. In contrast, our approach leverages the commercially available Pavlok device, which offers a range of feedback options -- including beeps, vibrations or shocks -- in adjustable nudging cycles. While Khan's system fulfils its intended functionality, its bulkiness and wired connections may cause discomfort for users during sleep -- an issue mitigated by the compact and wireless design of our system.

\begin{figure}[!t]
    \centering
    \includegraphics[width=0.6\linewidth]{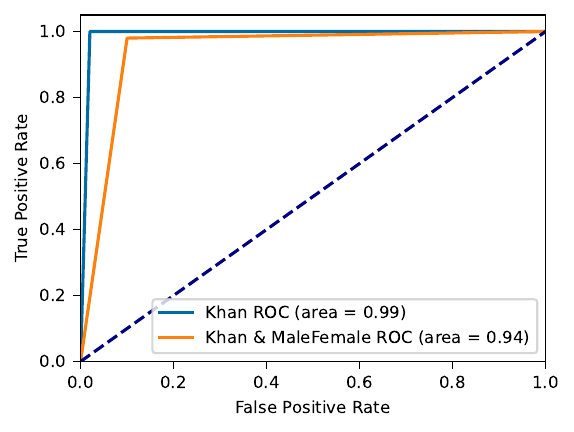}
    \caption{Receiver operating characteristic curve on the test sets.}
    \label{fig:roc}
\end{figure}

\Cref{fig:roc} depicts the receiver operating characteristic (ROC) curves for the test sets in two configurations: one trained on the Khan \cite{khan2019deep} dataset and the other trained on the combined Khan \cite{khan2019deep} \& MaleFemale \cite{oran2022malefemale} dataset. Although the model trained on the Khan dataset achieves an AUROC of 0.99, the model trained on the combined dataset achieves a slightly lower AUROC of 0.94. This difference suggests that the inclusion of additional data from the MaleFemale dataset may introduce complexities or variations that slightly affect the model's classification ability.

\subsection{Ablation Study}
\Cref{tab:model} presents the performance evaluation of different 1D convolutional layer configurations, varying in the number of output channels, when tested on the Khan \cite{khan2019deep} dataset. While three out of four configurations achieve a recall of 1.00, there are slight variations in accuracy, precision, F1 score and AUROC. The model with two 1D convolutional layers having output channels of 8 and 16 achieves the best performance. The smallest model, with two 1D convolutional layers (output channels of 4 and 8), has the fewest parameters (577) and achieves an accuracy of 0.96 and AUROC of 0.96. On the other hand, the largest model (1,873 parameters), with 16 output channels, yields an accuracy of 0.97, precision of 0.94, F1 score of 0.97 and AUROC of 0.97, slightly lower than the 8-16 configuration despite having more parameters. Since the 8-16 configuration strikes the best balance between model size and performance, we select this as the optimal configuration for our proposed system.

\begin{table}[!t]
    \caption{Performance comparison of 1D convolutional layer configurations on the Khan \cite{khan2019deep} dataset.}
    \label{tab:model}
    \centering
    \resizebox{\columnwidth}{!}{%
    \begin{tabular}{*7c} \toprule
    \rowcolor{mygray} \textbf{O/p Channels $\downarrow$} & \textbf{Params $\downarrow$} & \textbf{Accuracy $\uparrow$} & \textbf{Precision $\uparrow$} & \textbf{Recall $\uparrow$} & \textbf{F1 $\uparrow$} & \textbf{AUROC $\uparrow$} \\ \midrule
    8-16 & 1,337 & \textbf{0.99} & \textbf{0.98} & \textbf{1.00} & \textbf{0.99} & \textbf{0.99} \\
    4-8 & 577 & 0.96 & 0.93 & \textbf{1.00} & 0.96 & 0.96 \\
    8 & 945 & 0.95 & 0.94 & 0.96 & 0.95 & 0.95 \\
    16 & 1,873 & 0.97 & 0.94 & \textbf{1.00} & 0.97 & 0.97 \\ \bottomrule
    \end{tabular}%
    }
\end{table}


%

\subsection{Limitation}
While the neural network utilised in our proposed system for snoring detection provides accurate and reliable results, one of the limitations is that it is trained on general snoring clips and cannot differentiate between specific users. As a result, in scenarios where someone is sleeping next to their partner, the Pavlok device might be triggered to deliver a stimulus even if the user is not the one snoring. To tackle the problem, future work can integrate speaker diarisation~\cite{park2022review} to identify the person from whom the audio is coming in a multi-person scenario. The user interface can be revamped to make it more user-friendly. While the current study primarily focuses on the technical validation of the snore detection mechanism, empirical validation in practical settings lies beyond its scope. Future work shall include comprehensive user studies to evaluate the intervention's effectiveness and to explore user perceptions. We do not claim that this method can clinically or medically treat snoring. Instead, we present a tool that can be clinically tested to assist in the personal management of snoring.

\subsection{Ethical Use and Privacy Concerns}
Users are expected to respect the licenses of the deployed components used in Pavlok-Nudge. During run-time, Pavlok-Nudge detects and tracks audio signals in the user's living space. This could potentially breach the privacy of the subjects present in the environment. Thus, the user must obtain consent from the subjects before using our application. While developing this solution, we make sure that Pavlok-Nudge does not save any audio in the file system. Moreover, all the operations are performed in real-time; buffered audio signals are removed immediately after prediction.

\section{Conclusion} 
This paper proposes \textit{Pavlok-Nudge}, an interactive behaviour intervention technology (BIT) for atomic behaviour modification grounded in aversion therapy. At its core, a deep learning model detects the targeted behavioural pattern, such as snoring, and nudges the user through the Pavlok device. While snoring may seem like a minor inconvenience, it can have a significant impact on a person's health and well-being, or can be a sign of sleep apnea. One of the most difficult side effects of snoring is sleep disruption, which can lead to daytime fatigue and sleepiness. Our experiments on a public snoring benchmark result in the SOTA test performance using a lightweight 1D CNN architecture of only 1,337 trainable parameters. Apart from snoring, the Pavlok-Nudge framework has the potential to support other targeted behaviour modifications.

\section*{Acknowledgement}
This work was supported by resources provided by the Pawsey Supercomputing Research Centre with funding from the Australian Government and the Government of Western Australia. We thank Dr Susannah Soon, A/Professor at Curtin University, for her comments on an initial version of this paper.

%
%
%
\bibliographystyle{splncs04}
\bibliography{ref}
\end{document}